\documentclass[letterpaper, 10 pt, journal, twoside]{IEEEtran}

\usepackage{graphics}           
\usepackage{times}              
\usepackage{amsmath}            
\usepackage{amssymb}            
\usepackage{graphicx}
\usepackage{algorithm}
\usepackage[noend]{algpseudocode}
\usepackage{booktabs}
\usepackage{color}
\usepackage[nocompress]{cite} % sorts citations automatically, e.g., "[1],[2],[3]" instead of "[2],[3],[1]".
\definecolor{instructioncolor}{rgb}{.5,.5,.5}

\usepackage[font=small]{caption}

\def\eqref#1{Eq.~(\ref{#1})}

\makeatletter
\usepackage{xspace}
\DeclareRobustCommand\onedot{\futurelet\@let@token\@onedot}
\def\@onedot{\ifx\@let@token.\else.\null\fi\xspace}

\def\etal{{et al}\onedot}
\makeatother

\def\etalcite#1{\etal~\cite{#1}}

\usepackage{array}
\newcolumntype{L}[1]{>{\raggedright\let\newline\\\arraybackslash\hspace{0pt}}m{#1}}
\newcolumntype{C}[1]{>{\centering\let\newline\\\arraybackslash\hspace{0pt}}m{#1}}
\newcolumntype{R}[1]{>{\raggedleft\let\newline\\\arraybackslash\hspace{0pt}}m{#1}}

\usepackage{flushend}
\usepackage{dsfont}
\usepackage{graphicx}
\usepackage[colorlinks=true,
            citecolor=blue,
            linkcolor=blue,
            anchorcolor=blue,
            urlcolor=blue]{hyperref}
\usepackage{amsmath}
\usepackage{amssymb}
\usepackage{bbm}
\usepackage{multirow}
\usepackage{cite}
\usepackage{paralist}

\makeatletter
\renewcommand{\maketag@@@}[1]{\hbox{\m@th\normalsize\normalfont#1}}%
\makeatother

\title{Mobile-Seed: Joint Semantic Segmentation and Boundary Detection for Mobile Robots }

\author{
Youqi Liao, Shuhao Kang, Jianping Li, Yang Liu, Yun Liu, Zhen Dong, Bisheng Yang, 
Xieyuanli Chen  % <-this % stops a space
  \thanks{Manuscript received: Nov. 21, 2023; Revised: Jan. 15, 2024; Accepted: Feb. 20, 2024. This paper was recommended for publication by
  Editor Cesar Cadena Lerma upon evaluation of the Associate Editor and Reviewers' comments. Digital Object Identifier (DOI): see top of this page.}
  \thanks{
  This study was supported by the National Natural Science Foundation Project (No. 42201477, No. 42130105)~(Corresponding author: Jianping Li)} %
  \thanks{Y. Liao, Z. Dong and B. Yang are with Wuhan University, China. S. Kang is with the Technical University of Munich, Germany. J. Li is with Wuhan University, China and Nanyang Technological University, Singapore. Yang Liu is with the King’s College London, UK.  Yun Liu is with the Institute of Infocomm Research (I2R), A*STAR, and X. Chen is with the National University of Defense Technology, China. }%
}

\begin{document}
\maketitle

\markboth{IEEE Robotics and Automation Letters.}{Liao \MakeLowercase{\textit{et al.}}: Mobile-Seed: Joint Semantic Segmentation and Boundary Detection for Mobile Robots}

%%%%%%%%%%%%%%%%%%%%%%%%%%%%%%%%%%%%%%%%%%%%%%%%%%%%%%%%%%%%%%%%%%%%%%%%%%%%%%%%
\begin{abstract}
 Precise and rapid delineation of sharp boundaries and robust semantics is essential for numerous downstream robotic tasks, such as robot grasping and manipulation, real-time semantic mapping, and online sensor calibration performed on edge computing units. Although boundary detection and semantic segmentation are complementary tasks, most studies focus on lightweight models for semantic segmentation but overlook the critical role of boundary detection. In this work, we introduce Mobile-Seed, a lightweight, dual-task framework tailored for simultaneous semantic segmentation and boundary detection.  Our framework features a two-stream encoder, an active fusion decoder (AFD) and a dual-task regularization approach. The encoder is divided into two pathways: one captures category-aware semantic information, while the other discerns boundaries from multi-scale features. The AFD module dynamically adapts the fusion of semantic and boundary information by learning channel-wise relationships, allowing for precise weight assignment of each channel. Furthermore, we introduce a regularization loss to mitigate the conflicts in dual-task learning and deep diversity supervision. Compared to existing methods, the proposed Mobile-Seed offers a lightweight framework to simultaneously improve semantic segmentation performance and accurately locate object boundaries. Experiments on the Cityscapes dataset have shown that Mobile-Seed achieves notable improvement over the state-of-the-art (SOTA) baseline by 2.2 percentage points (pp) in mIoU and 4.2 pp in mF-score, while maintaining an online inference speed of 23.9 frames-per-second (FPS) with 1024$\boldsymbol{\times}$2048 resolution input on an RTX 2080 Ti GPU. Additional experiments on CamVid and PASCAL Context datasets confirm our method's generalizability. Code and additional results are publicly available at: \url{https://whu-usi3dv.github.io/Mobile-Seed/}.
 \end{abstract}
 

%%%%%%%%%%%%%%%%%%%%%%%%%%%%%%%%%%%%%%%%%%%%%%%%%%%%%%%%%%%%%%%%%%%%%%%%%%%%%%%%
\section{Introduction}
\label{sec:intro}

\begin{figure}[ht]
  \centering
  \includegraphics[width=0.95\linewidth]{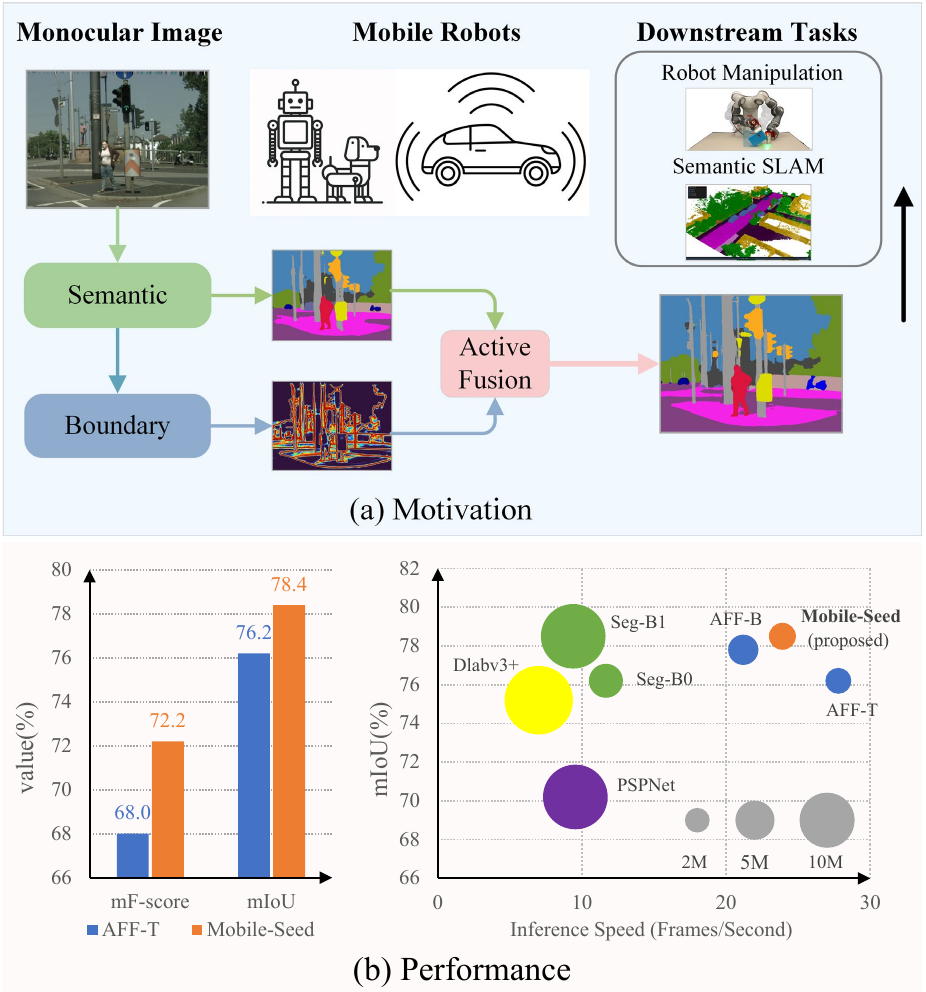}
  \vspace{-0.1cm}
  \caption{(a) Motivation map: Mobile-Seed performs pixel-wise segmentation and object boundary detection simultaneously, and then fuses semantic and boundary features for accurate prediction. The boundary detection and semantic segmentation predictions could be transferred for downstream tasks, e.g., robot manipulation, semantic mapping and sensor calibration. (b) Our Mobile-Seed achieves higher performance on both semantic segmentation and boundary detection tasks while keeping real-time efficiency. The resolution of input is 1024$\times$2048 when testing inference speed. ``AFF" and ``Seg" mean the AFFormer\cite{dong2023head} and SegFormer\cite{xie2021segformer}, respectively.} 
  \label{fig:motivation}
  \vspace{-0.2cm}
\end{figure}

Semantic segmentation and boundary detection are fundamental tasks for simultaneous localization and mapping (SLAM)~\cite{chen2020overlapnet}, autonomous driving ~\cite{2022SLidR}, behavior prediction~\cite{gao2020vectornet} and sensors calibration~\cite{liao2023se}. Semantic segmentation predicts the categorical labels for each pixel, and the boundary detection task identifies pixels lying on the boundary area where at least one neighborhood pixel belongs to a different class. As the boundary always surrounds the object's body~\cite{zhen2020joint}, robust prediction of the body label guides the object boundary detection, while improving the boundary location is crucial for semantic segmentation accuracy. In other words, semantic segmentation and boundary detection are complementary tasks. Moreover, simultaneously extracting segmentation and boundary information in compact robotics is important for semantic SLAM~\cite{zhang2014loam,chen2019suma++}, in which the boundary is a strong constraint for solving the relative pose and location, and segmentation is crucial for dynamic object removal. However, on the one hand, most lightweight approaches~\cite{xie2021segformer,zhang2022topformer,dong2023head} attempt to solve the semantic segmentation task but overlook the boundary accuracy. On the other hand, existing dual-task learning approaches~\cite{takikawa2019gated,li2020improving,zhen2020joint} design novel architectures for performance improvement but neglect the computational burden. Overall, simultaneously capturing the segmentation and boundary has not received enough attention, but this is precisely in urgent need of the robotics society. In this paper, we investigate how to design a lightweight framework for jointly learning the semantic and boundary mask in a complementary
manner, as shown in Fig.~\ref{fig:motivation}.

\begin{figure}[t]
  \centering
  \includegraphics[width=0.9\linewidth]{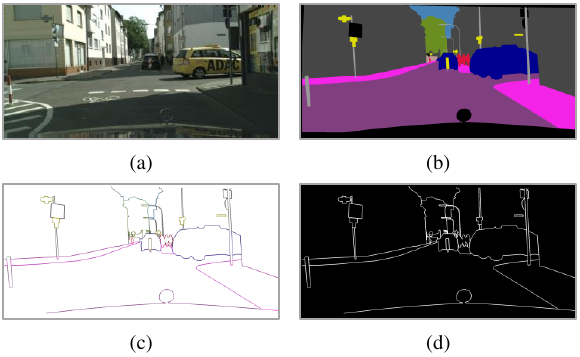}
  \vspace{-0.2cm}
  \caption{Example diagram of color image (a), semantic mask (b), semantic boundary mask (c) and binary boundary mask (d). Semantic boundary masks are generated as \cite{yu2017casenet,hu2019dynamic}, and binary boundary masks are generated as \cite{liu2022semantic}.}
  \label{fig:example}
  \vspace{-0.2cm}
\end{figure}

To deploy semantic segmentation for real-world online robotic and autonomous driving applications, powerful yet lightweight vision transformers (ViT)~\cite{dosovitskiy2020image} have been developed. 
For example, the hierarchical attention~\cite{xie2021segformer}, stride attention~\cite{zhang2022topformer}, and window attention~\cite{mehta2021mobilevit} are proposed to capture the long-range context with low computation cost and outperform convolution neural network (CNN) based methods by a large margin.
However, these advances are still insufficient to accurately locate object boundaries. The main reasons are: i) as the Transformer lacks inductive bias \cite{dosovitskiy2020image}, it is not good at capturing fine-grained details in a local window; ii) most methods adopt very simple decoder designs, which lack the ability to capture and recover details. Some recent approaches~\cite{dong2023head,wan2023seaformer} even remove the decoder for efficiency, called ``head free'', which exacerbates boundary blurring. For the boundary detection task, most existing approaches~\cite{yu2017casenet,hu2019dynamic,liu2022semantic} overlooked the computational efficiency. In the field of dual-task learning for semantic segmentation and boundary detection, several approaches~\cite{takikawa2019gated,li2020improving,xiao2023baseg} pointed out that jointly learning the boundary detection and semantic segmentation tasks with reasonable designs benefits both tasks, but none of them discussed how to implement with a lightweight design for mobile robots. It should be retained that the boundary detection task here is significantly different from the edge detection task~\cite{liu2017richer}. Fig.~\ref{fig:example} shows the semantic mask, the semantic boundary mask, and the binary boundary mask of a color image. Boundary detection aims to find semantically discontinuous areas instead of dramatic intensity, illumination, or texture changes in edge detection task. 

\begin{figure*}[ht]
  \centering
  \includegraphics[width=0.9\linewidth]{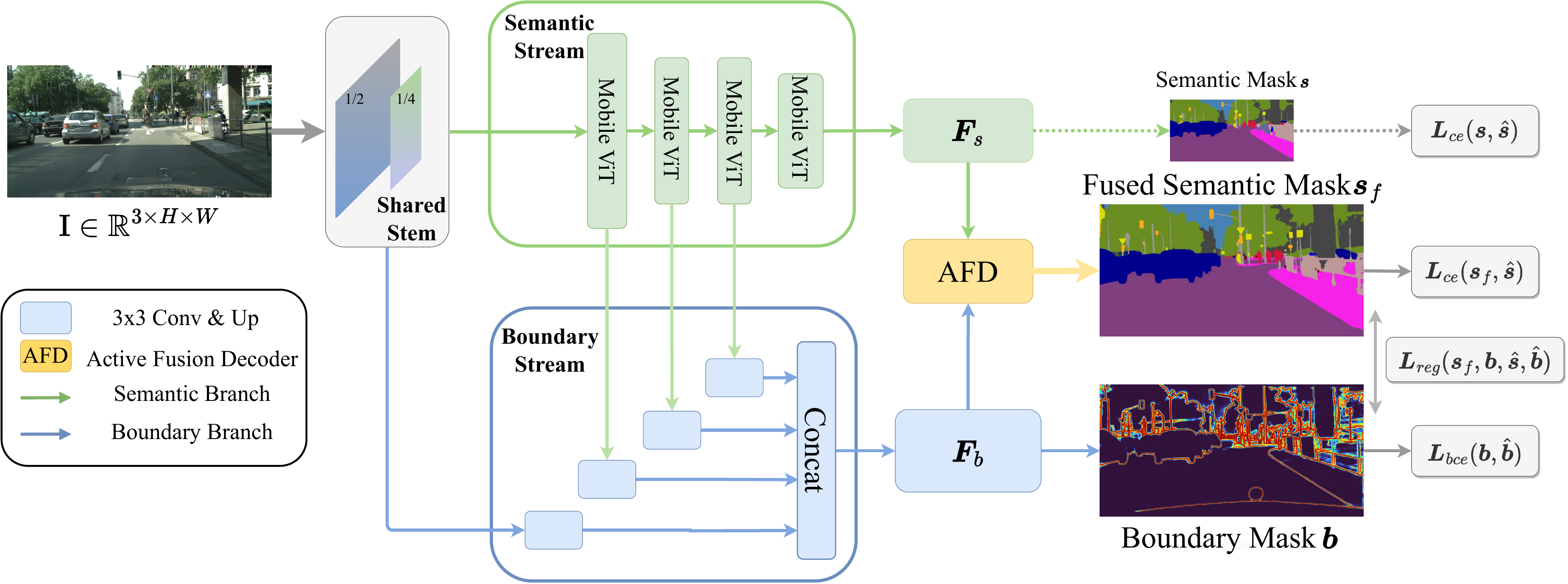}
  \caption{Workflow of Mobile-Seed, where the semantic stream $\mathcal{S}$ and boundary stream $\mathcal{B}$ extract semantic and boundary features respectively. AFD estimates the relative weights for each channel of semantic features $\boldsymbol{F}_s$ and boundary features $\boldsymbol{F}_b$. An auxiliary classification head is applied to the semantic stream for direct supervision during training. Semantic prediction $\boldsymbol{s}$, fused semantic prediction $\boldsymbol{s}_f$, and boundary prediction $\boldsymbol{b}$ are supervised separately and accordingly. Regularization loss $\mathcal{L}_{reg}$ mitigates the divergences caused by dual-task learning.}
  \label{fig:workflow}
  \vspace{-0.2cm}
\end{figure*}

To address the limitations mentioned above, we present a lightweight framework for simultaneous semantic segmentation and boundary detection. The workflow is shown in Fig.~\ref{fig:workflow}. Our objective is to utilize the semantic stream to offer fundamental knowledge for the boundary stream while supplementing the semantic segmentation task with fine-grained details captured by the boundary branch. Additionally, we introduce the active fusion decoder (AFD) to learn the fusion weights from inputs and fuse the semantic and boundary features in a dynamic way. Furthermore, we incorporate the dual-task regularization losses to alleviate conflicts arising from deep diverse supervision (DDS)~\cite{liu2022semantic}. Experiments on multiple public datasets demonstrate our Mobile-Seed outperforms existing methods by a large margin, especially in predicting crisper boundaries and segmenting small and thin objects. Overall, the main contributions of this paper include:
 \begin{enumerate}
     \item We propose a lightweight joint semantic segmentation and boundary detection framework for mobile robots. This framework can concurrently learn both the boundary mask and semantic mask. 
     \item We present the AFD for learning the channel-wise relationship between semantic features and boundary features. Compared to the fixed weight methods (fusion weights independent of the input), our AFD is more flexible in assigning proper weights for semantic features and boundary features. 
     \item We introduce the dual-task regularization loss to effectively mitigate conflicts arising from DDS, allowing the tasks of semantic segmentation and boundary detection to contribute to each other.
 \end{enumerate}
 
%%%%%%%%%%%%%%%%%%%%%%%%%%%%%%%%%%%%%%%%%%%%%%%%%%%%%%%%%%%%%%%%%%%%%%%%%%%%%%%%
\section{Related Work}\label{sec:2}

\subsection{Lightweight Semantic Segmentaiton}
Since the pioneering approaches fully convolution network (FCN)~\cite{long2015fully} ushered in a new era, a significant amount of works~\cite{zhao2017pyramid,chen2018encoder} have been dedicated to addressing semantic segmentation tasks. To reduce the computational burden caused by dense convolution operations on feature maps, the MobileNet~\cite{howard2017mobilenets} proposed the depthwise separable convolution and ShuffleNet~\cite{zhang2018shufflenet} proposed the channel shuffle to maintain accuracy. Fast-SCNN~\cite{poudel2019fast} proposed the ``learning to downsample'' module to produce shared low-level fundamental features. During the transformer era, SegFormer~\cite{xie2021segformer} was the first transformer-based lightweight design for mobile devices. Activated by Swin-Transformer~\cite{liu2021swin}, MobileViT~\cite{mehta2021mobilevit} proposed the hybrid CNN and Transformer blocks for local and global processing. TopFormer~\cite{zhang2022topformer} designs the token pyramid module for scale-aware features. Experiments show that Topformer achieves a better trade-off between accuracy and efficiency than previous approaches. AFFormer~\cite{dong2023head} proposed the channel-wise attention module and SeaFormer~\cite{wan2023seaformer} proposed the axial-attention module. Coincidentally, both of them utilized the ``head free'' decoder design: a simple classification head with several convolution layers, which means predicting at a low resolution without progressive upsampling and refinement. PP-mobileSeg~\cite{tang2023pp} inherited the stride-Former frame~\cite{wan2023seaformer} and proposed the aggregated attention module (AAM) and valid interpolation module (VIM) to enhance the semantic features. Unlike the above approaches which design single-branch models for semantic segmentation, we introduce a dual-branch framework to simultaneously learn semantic segmentation and boundary detection.

\subsection{Boundary Detection}
CASENet~\cite{yu2017casenet} is the first multi-label learning framework to identify semantic boundaries. Based on the ResNet-101~\cite{he2016deep}, CASENet utilizes the bottom layers for details and the top layers for category-aware features. STEAL~\cite{acuna2019devil} detects semantic boundaries and corrects noise labels iteratively for crisper prediction. DFF~\cite{hu2019dynamic} proposed to learn dynamic weights for different input images and locations. RPCNet~\cite{zhen2020joint} proposed to jointly learn the semantic and semantic boundary with iterative pyramid context modules. DDS~\cite{liu2022semantic} proposed the information converter consisting of several ResNet blocks~\cite{he2016deep} to mitigate the conflicts caused by deep diverse supervision. However, integrating the information converter into the network will significantly increase the computational burden, especially for high-resolution images.

GSCNN~\cite{takikawa2019gated}, DecoupleNet~\cite{li2020improving} and BASeg~\cite{xiao2023baseg} are the most similar approaches to our work. They take binary boundary detection as a supplement for semantic segmentation, which is performed in an auxiliary manner like the auxiliary loss function. However, our approach has significant differences compared to them: (i) Mobile-Seed is a joint boundary detection and semantic segmentation framework instead of a boundary-auxiliary semantic segmentation; (ii) we focus on designing a lightweight model with the least computational burden in contrast to previous cumbersome models.

%%%%%%%%%%%%%%%%%%%%%%%%%%%%%%%%%%%%%%%%%%%%%%%%%%%%%%%%%%%%%%%%%%%%%%%%%%%%%%%%
\section{Mobile-Seed Overview}\label{sec:3}
In this section, we present the lightweight  Mobile-Seed for joint semantic segmentation and boundary detection learning. As illustrated in Fig.~\ref{fig:workflow}, the Mobile-Seed contains a two-stream encoder for semantic segmentation and boundary detection, and then an active fusion decoder (AFD) for features fusion. Each branch's output is supervised with the corresponding ground truth. Moreover, regularization loss is introduced to direct dual-task learning in a complementary way.

\subsection{Architecture Overview} \label{sec:arc}

Since the goal is to learn the semantic and boundary information simultaneously, we propose a two-stream encoder to capture the corresponding features from the input image. Firstly, a simple shared stem module consisting of two MobileNetV2 blocks~\cite{sandler2018mobilenetv2} is utilized to embed the original image $\mathrm{I}\in \mathbb{R}^{3\times H \times W}$ into high-dimension feature space, where $H$ and $W$ mean the height and width of image $\mathrm{I}$ respectively. The semantic stream $\mathcal{S}$ takes the second embed feature map as input and generates semantic-rich features. We emphasize that the semantic stream could be any lightweight semantic segmentation backbone, e.g.,~\cite{howard2017mobilenets,zhang2018shufflenet,xie2021segformer,zhang2022topformer,xu2023pidnet,wan2023seaformer,dong2023head}. In this paper, we select one of the most recent SOTA methods, AFFormer-T~\cite{dong2023head} (`T' means the tiny model of AFFormer) as our semantic stream backbone. A simple classification head is used to generate the auxiliary semantic map $\boldsymbol{s}$ during training.

\begin{figure}[t]
  \centering
  \includegraphics[width=0.9\linewidth]{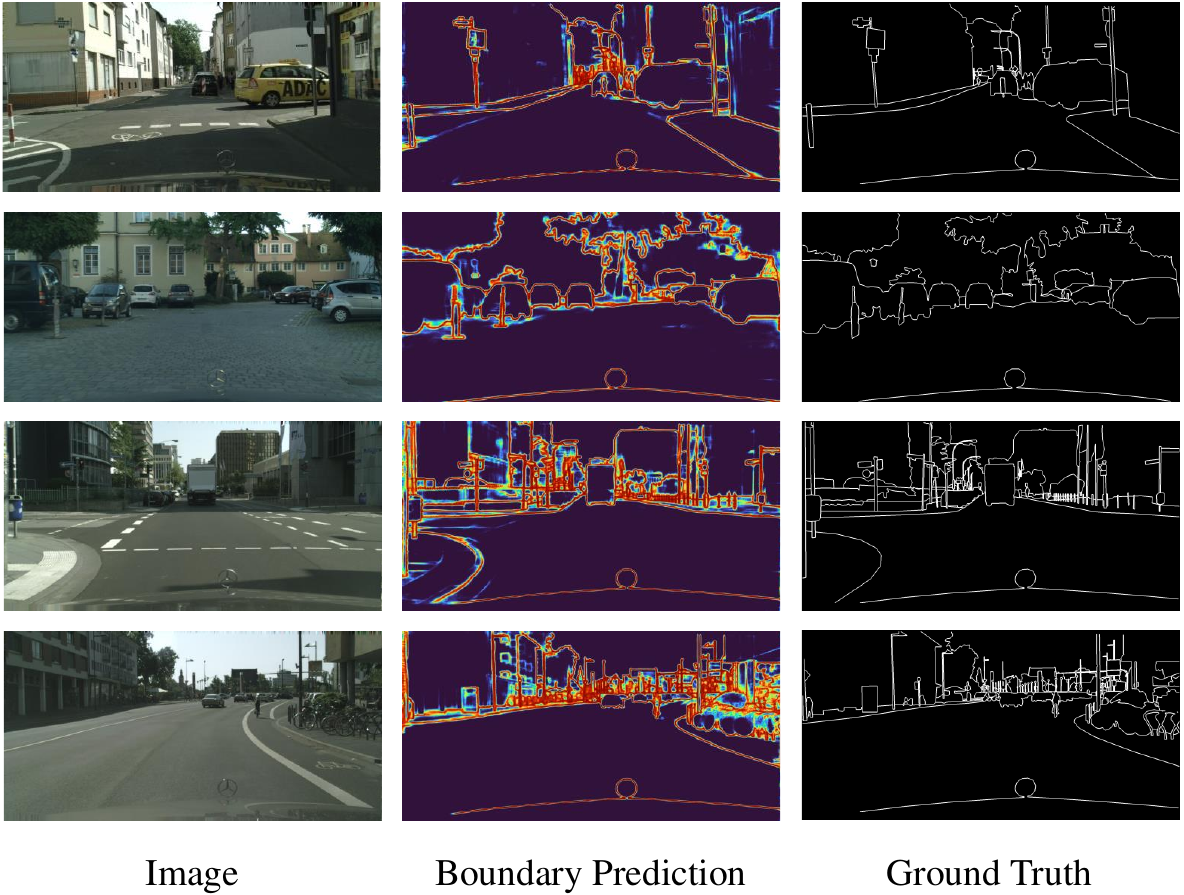}
  \vspace{-0.2cm}
  \caption{Examples of boundary maps from the boundary stream. The first column shows the input, the second shows the boundary predictions, and the last column shows the ground-truth boundaries.}
  \label{fig:binary_edge}
\end{figure}

The boundary stream $\mathcal{B}$ takes the first embedded feature map and intermediate feature maps of the semantic stream as input, and feeds into a $3\times3$ convolution layer, group normalization layer and ReLU layer to differentiate the semantic features to boundary features. Let $m$ denote the stage number and $i \in \{1,2,...,m\}$ denote the running index, the $i$-th stage's representation of the semantic stream is denoted as $\boldsymbol{F}_s^i$. For the $i$-th location, the information conversion process in the boundary stream is denoted as:
\begin{equation}
    \boldsymbol{F}_b^i = \sigma(C_{3\times3}(\boldsymbol{F}_s^i)),
\end{equation}
where $\boldsymbol{F}_b^i$ means boundary feature of the $i$-th stage, $C_{3\times3}$ means normalized $3\times3$ convolution layer, and $\sigma$ means activation operation. Then, the multi-scale boundary features are upsampled with bilinear interpolation and concatenated together. Finally, a simple classification head is applied for predicting boundary map $\boldsymbol{b}\in \mathbb{R}^{H \times W}$, as shown in Fig.~\ref{fig:binary_edge}. 

\subsection{Active fusion Decoder}  \label{sec:decoder}
After obtaining high-dimensional semantic and boundary features, the ensuing problem is how to efficiently fuse features from different domains. As the semantic stream is supervised to learn category-aware semantics and the boundary stream is supervised to learn category-agnostic boundaries, there is a significant domain divergence in two types of features. Most previous approaches use addition~\cite{long2015fully,zhang2022topformer} or concatenation~\cite{li2020improving} to fuse features from multiple scales or streams, and some others introduce atrous spatial pyramid pooling (ASPP)~\cite{takikawa2019gated} or pyramid context module (PCM)~\cite{zhen2020joint} for well-mixed in spatial dimension. The above methods could be classified as fixed weights methods, where the fusion weights in the channel dimension are image-independent. However, the importance of each channel in semantic features and boundary features may vary for different images. Therefore, the fusion weights should be conditioned on the input. There are dynamic fusion methods~\cite{hu2019dynamic,xiao2023baseg} that can adapt the weights for semantic edge detection and semantic segmentation tasks. However, calculating fusion weights in both spatial and channel dimensions is still too cumbersome for the lightweight framework.

\begin{figure}[t]
  \centering
  \includegraphics[width=0.9\linewidth]{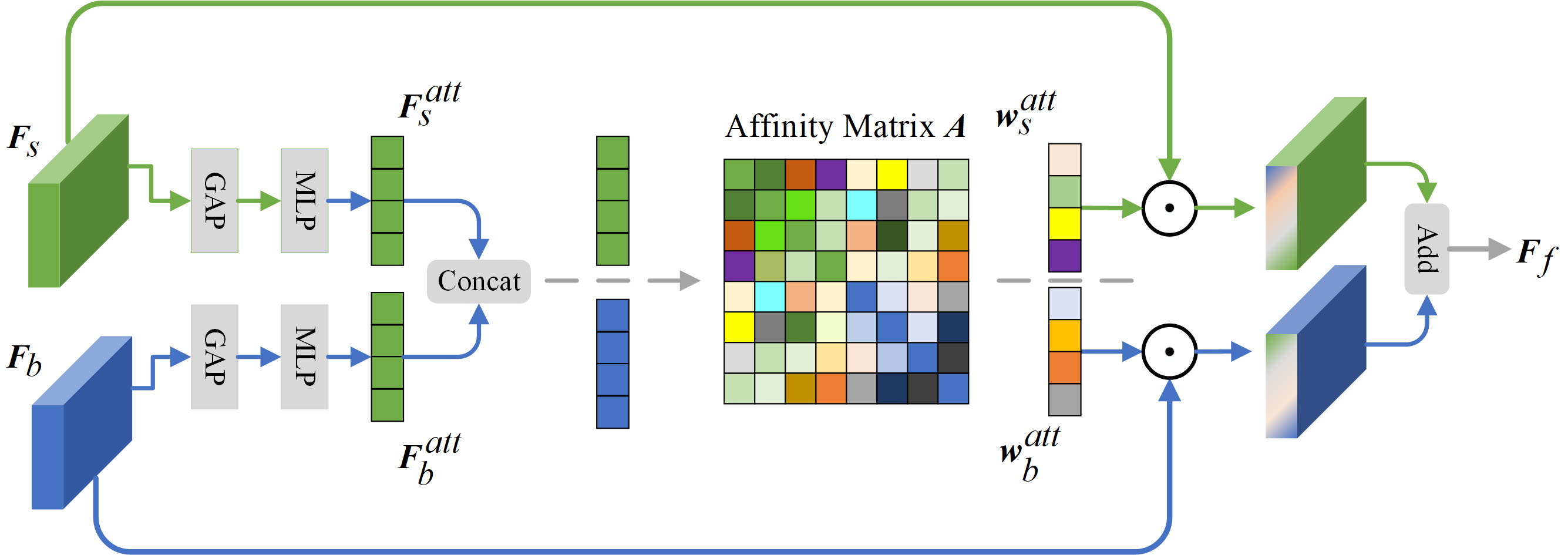}
  \caption{Illustration of the proposed AFD.}
  \label{fig:AFD}
  \vspace{-0.2cm}
\end{figure}

To tackle this issue, we propose the active fusion decoder (AFD) module to learn the fusion weights for semantic stream and boundary stream, as shown in Fig.~\ref{fig:AFD}. With the semantic feature map $\boldsymbol{F}_{s} \in \mathbb{R}^{C \times H \times W}$ from semantic stream and boundary feature map $\boldsymbol{F}_{b} \in \mathbb{R}^{C \times H \times W}$ from boundary stream, we first calculate the semantic channel-wise attention vector $\boldsymbol{F}_s^{att} \in  \mathbb{R}^{C \times 1 \times 1}$ and boundary channel-wise attention vector $\boldsymbol{F}_b^{att} \in  \mathbb{R}^{C \times 1 \times 1}$ with global average pooling (GAP):
\begin{equation}\label{eq:gap}
\begin{split}
     & \boldsymbol{F}_s^{att}=f_s(GAP(\boldsymbol{F}_s)), \\
     & \boldsymbol{F}_b^{att}=f_b(GAP(\boldsymbol{F}_b)),
\end{split}    
\end{equation}
where $\boldsymbol{F}_s^{att},\boldsymbol{F}_b^{att}\in \mathbb{R}^{C \times 1 \times 1}$, and $f(\cdot)$ means multi-layer perception (MLP) modules. We stack attention vectors of semantic and boundary features in channel dimension:
\begin{equation}\label{eq:cat}
    \boldsymbol{F}_{f}^{att} = (\boldsymbol{F}_{s}^{att} \| \boldsymbol{F}_{b}^{att}),
\end{equation}
where $\|$ means channel-wise concatenation operation. To metric the affinity among each channel of semantic and boundary features, we split the fusing attention vector  $\boldsymbol{F}_{f}^{att} \in \mathbb{R}^{2C}$ into $H$ groups and generate the channel-wise affinity matrix $\boldsymbol{A}\in \mathbb{R}^{2C/H \times 2C/H}$. We first calculate the query vector $\boldsymbol{q}$, key vector $\boldsymbol{k}$ and value vector $\boldsymbol{v}$ from $\boldsymbol{F}_{f}^{att}$ by linear projection, and then each component of affinity matrix $\boldsymbol{A}_{i,j}$ is computed as:
\begin{equation}
    \boldsymbol{A}_{i,j} = e^{\boldsymbol{q}_i{\boldsymbol{k}_j}^\top}/\sum_{i=1}^{H}{e^{\boldsymbol{q}_i}},
\end{equation}
where $\boldsymbol{q}_i$ is the $i$-th head of query, and $\boldsymbol{k}_j$ is the $j$-th head of key. The $i$-th head of active fusion weight $\boldsymbol{w}$ is calculated with:
\begin{equation}\label{eq:weight}
    \boldsymbol{w}_i =  \boldsymbol{A} \boldsymbol{v}_i,
\end{equation}
where $\boldsymbol{v}_i$ means the $i$-th head of value vector $\boldsymbol{v}$. Then we divide the weight vector $\boldsymbol{w}$ into $\boldsymbol{w}_{s}$ for semantic stream and $\boldsymbol{w}_{b}$ for boundary stream by inverse operation of Eq.~\ref{eq:cat}. The fused features $\boldsymbol{F}_{f}$ is calculated through residual connection:
\begin{equation}\label{eq:fuse}
    \boldsymbol{F}_{f} = (1 + \boldsymbol{w}_{s})  \boldsymbol{F}_s + (1 + \boldsymbol{w}_b)  \boldsymbol{F}_{b}.
\end{equation}
A $1\times1$ convolution layer is followed as classification head to compact the $\boldsymbol{F}_f$ into the final semantic prediction map $\boldsymbol{s}_f \in \mathbb{R}^{N \times H \times W}$, $N$ means the category number. Overall, the AFD estimates the channel-wise relationship within and among semantic features and boundary features for learning optimal fusion weights.

\subsection{Loss Functions}
In the Mobile-Seed framework, we jointly train the semantic and boundary stream in an end-to-end way and use the AFD to fuse the dual-task features for final prediction. Therefore, available supervisions include semantic label $\hat{\boldsymbol{s}} \in \mathbb{R}^{N \times H\times W}$, semantic boundary label $\hat{\boldsymbol{b}_s} \in \mathbb{R}^{N \times H\times W}$ and binary boundary label $\hat{\boldsymbol{b}} \in \mathbb{R}^{ H\times W}$ (as shown in Fig.~\ref{fig:example} (b), (c) and (d)). The cross-entropy (CE) loss and binary cross-entropy (BCE) loss function are used to supervise the semantic and boundary predictions:
\begin{equation}\label{eq:loss_base}
\small
    \mathcal{L}_{cls}\,=\, \mathcal{L}_{ce}(\boldsymbol{s},\hat{\boldsymbol{s}})\,+\, \mathcal{L}_{ce}(\boldsymbol{s}_f,\hat{\boldsymbol{s}})\,+\, \mathcal{L}_{bce}(\boldsymbol{b},\hat{\boldsymbol{b}}).
\end{equation}

\textbf{Dual-task regularization.} With the top supervision of semantic label $\hat{\boldsymbol{s}}$, the top layers are acquired to learn abstracted semantic representation, enabling it to cover diverse object shapes, lighting conditions and textures. In contrast, bottom supervision of the boundary label $\hat{\boldsymbol{b}}$ leads the bottom layers to distinct boundaries or non-boundaries, rather than category-aware semantics. Since the bottom layers provide basic representations for both semantic segmentation and boundary detection, the bottom layers receive two distinctively different supervisions under back-propagation. Liu~\etalcite{liu2022semantic} pointed out that applying deep diverse supervision (DDS) directly may lead to conflicts and performance degradation, while a single convolution layer in the boundary stream is too weak to alleviate the supervision conflicts. Our ablation studies in Sec.~\ref{sec:ablation} also support this finding. To address these conflicts, authors of the DDS introduced buffer blocks to isolate the backbone and side layers. Unlike that, we design bi-directional consistency loss $\mathcal{L}_{reg}$ consisting of the semantic-to-boundary consistency loss $\mathcal{L}_{s2b}$ and the boundary-to-semantic consistency loss $\mathcal{L}_{b2s}$ to soften the conflict and free computation burden during inference.
The semantic-to-boundary consistency loss $\mathcal{L}_{s2b}$ is designed to align pseudo semantic boundary $\boldsymbol{b}_{ps}$ generated from semantic prediction $\boldsymbol{s}_f$ with the semantic boundary label $\hat{\boldsymbol{b}}_s$. We introduce a filtering template $\boldsymbol{T} \in \mathbb{R}^{N \times (2r+1)\times (2r+1)}$ to look into neighbors of each pixel and seek for the maximum difference in each category, where $r$ is the search radius. For ease of description, we set $r=2$ here and each channel of the filtering template $\boldsymbol{T}$ is:
\begin{equation}
 \boldsymbol{T}_i = \begin{bmatrix}
    0 & 0 & 1 & 0 & 0 \\
    0 & 1 & 1 & 1 & 0 \\
    1 & 1 & -1 & 1 & 1 \\
    0 & 1 & 1 & 1 & 0 \\
    0 & 0 & 1 & 0 & 0 
\end{bmatrix}, i \in 1,2,...,N.   
\end{equation}
 We slide the template on the fused semantic prediction map $\boldsymbol{s}_{f} \in \mathbb{R}^{N \times H \times W}$ and select the max absolute value in the filtering window as the pseudo semantic boundary prediction:
\begin{equation}\label{eq:pe}
    \boldsymbol{b}_{ps}=\max_{\boldsymbol{T}}(\| \boldsymbol{T} \circledast \boldsymbol{s}_f \|).
\end{equation}
Intuitively, the template $\boldsymbol{T}$ mimics the generation process of the semantic boundary label $\hat{\boldsymbol{b}}_s$ by checking whether neighboring pixels have different labels or not. Mean absolute loss is used to supervise the pseudo semantic boundaries: 
\begin{equation}
    \mathcal{L}_{s2b} = \| \boldsymbol{b}_{ps} - \hat{\boldsymbol{b}_s} \|.
\end{equation}

On the other hand, boundary prediction provides important prior knowledge to ensure semantic consistency between body and boundary. We combine weighted cross-entropy loss with boundary prior to formulating the boundary-to-semantic consistency loss $\mathcal{L}_{b2s}$ :
\begin{equation}
    \mathcal{L}_{b2s} = \sum_{k,p} \mathds{1}_{\boldsymbol{b},\hat{\boldsymbol{b}}} \big(\hat{\boldsymbol{s}}(k,p) \log (\boldsymbol{s}_f(k,p)\big),
\end{equation}
where $k$ and $p$ walk over the categories and pixels. $\mathds{1}_{\boldsymbol{b},\hat{\boldsymbol{b}}}=\{1: \boldsymbol{b} > \epsilon \cup \hat{\boldsymbol{b}} = 1\}$ marks ground-truth (GT) pixels and high confidence pixels on the boundary prediction map $\boldsymbol{b}$. $\epsilon$ is the confidence threshold and we use 0.8 in the experiments. With the bi-direction consistency losses $\mathcal{L}_{s2b}$ and $\mathcal{L}_{b2s}$, the regularization function can be formulated as :
\begin{equation}
    \mathcal{L}_{reg}\, = \, \mathcal{L}_{s2b}\, +\, \mathcal{L}_{b2s}.
\end{equation}
The total loss function is:
\begin{equation}
\begin{split}
     \mathcal{L}\,  = \, \lambda_1\mathcal{L}_{cls}\, +\, \lambda_2\mathcal{L}_{reg},
\end{split}
\end{equation}
where $\lambda_1, \lambda_2$ are hyper-parameters to control weights of classification loss and dual-task regularization.

%%%%%%%%%%%%%%%%%%%%%%%%%%%%%%%%%%%%%%%%%%%%%%%%%%%%%%%%%%%%%%%%%%%%%%%%%%%%%%%%
\section{Experimental Evaluation}\label{sec:4}
\label{sec:exp}

In this section, we conduct experiments on three publicly available datasets: Cityscapes \cite{cordts2016cityscapes}, CamVid \cite{brostow2009semantic} and PASCAL Context \cite{mottaghi2014role} to show the capability of our method in various environments. Sec.~\ref{sec:setup} introduces the implementation details and evaluation metrics. In Sec.~\ref{sec:results}, we compare our method with SOTA methods on the Cityscapes dataset and extensively validate on CamVid and PASCAL Context to demonstrate the generalization and robustness. In Sec.~\ref{sec:ablation}, ablation studies on the AFD and regularization loss demonstrate the effectiveness of our design. Overall, the results prove that our approach could (i) jointly learn the semantic segmentation and boundary detection tasks; (ii) improve the semantic segmentation performance while maintaining online operation; (iii) accurately detect object boundaries in complex scenes.

\subsection{Experimental Setup}\label{sec:setup}
\textbf{Implementation details.}
We build our model based on the MMsegmentation toolbox. All experiments were performed on an NVIDIA RTX 4090 GPU. We select the AFFormer-T~\cite{dong2023head} as our semantic stream and pre-train on ImageNet-1k~\cite{deng2009imagenet}, while the boundary stream learns from scratch. Most training details follow previous approaches~\cite{dong2023head,zhang2022topformer}. 
The hyperparameters for controling loss weight are set to $\lambda_1=\lambda_2=1$. We use the AdamW optimizer~\cite{loshchilov2017decoupled} for all datasets to update model parameters. The data augmentation methods include random resize, random scaling, random horizontal flipping and color jittering. The training iterations, batch size and input image size for Cityscapes, CamVid and PASCAL Context datasets are set to [160K, 8, 1024$\times$1024], [20K, 16, 520$\times$520], [80K, 16, 480$\times$480] respectively. For more training details, please refer to our open-source code.

\textbf{Evaluation metrics.} We report three quantitative measures to evaluate the performance of our method. (i) We evaluate the semantic segmentation results with the widely used Intersection over Union (IoU) metric. (ii) To evaluate our purpose that the Mobile-Seed extracts high-quality semantic boundary, we use the F-score as previous approaches \cite{takikawa2019gated,li2020improving} on the Cityscapes \textit{val} dataset. This boundary metric computes the F1 score between dilated semantic boundary prediction and ground truth with a threshold to control the bias degree. We set thresholds 0.00088, 0.001875, 0.00375, and 0.005, which correspond to 3, 5, 9, and 12 pixels respectively. (iii) The boundary IoU (BIoU) \cite{cheng2021boundary} is introduced to further evaluate both the semantic boundary and binary boundary performance on various datasets. Compared with the F-score, the BIoU is more sensitive to small object errors. For efficiency analysis, we report the FLOPs, params number and FPS evaluated on an RTX 2080 Ti GPU with batch size of 1. For a fair comparison, inferences are conducted on the origin image resolution instead of multi-scale inference.

%%%%%%%%%%%%%%%%%%%%%%%%
\subsection{Quantatitave and Qualitative Results}\label{sec:results}
\begin{figure}[t]
  \vspace{-0.2cm}
  \centering
  \includegraphics[width=0.9\linewidth]{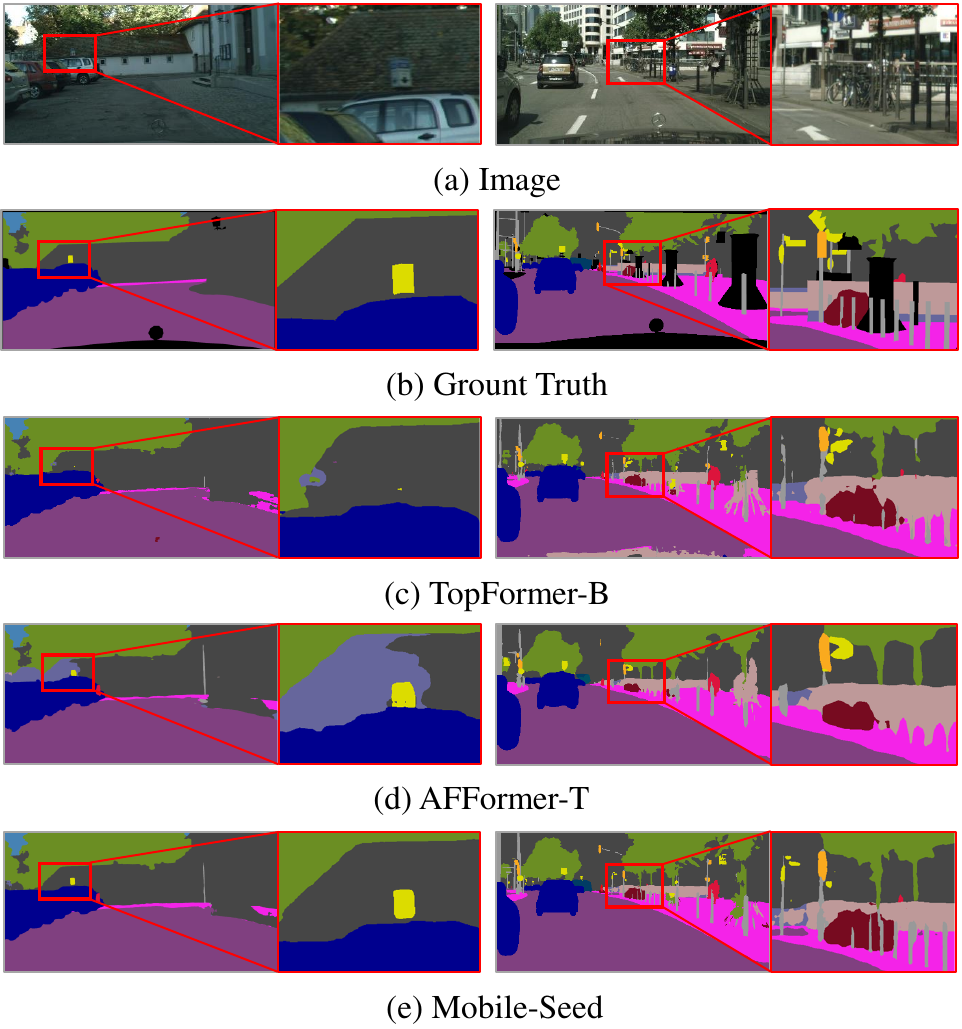}
  \vspace{-0.2cm}
  \caption{Qualitative comparisons for semantic segmentation. The ``unlabeled'' area is rendered as black in ground truth.}
  \label{fig:qualitative}
  \vspace{-0.2cm}
\end{figure}

The comparison of the semantic segmentation results with SOTA methods on the Cityscapes \textit{val} dataset is shown in Tab.~\ref{tab:table1}. As can be seen, the Mobile-Seed owns fewer parameters, lower computation costs, and higher mIoU performance than AFFormer-B, validating that our two-stream design achieves a better balance of accuracy and efficiency. Tab.~\ref{tab:table2} shows the category-wise comparison in terms of IoU with our strong baseline method AFFormer-T. Our method significantly outperforms the baseline method in most categories (18/19), improving the mIoU score from 76.2 to 78.4 ($2.2\%$ improvement) over the strong baseline. Moreover, our method could still keep near real-time (23.9 FPS) inference speed. The qualitative results are shown in Fig.~\ref{fig:qualitative}, with additional results available on the project page.

\begin{figure*}[ht]
  \centering
  \includegraphics[width=0.9\linewidth]{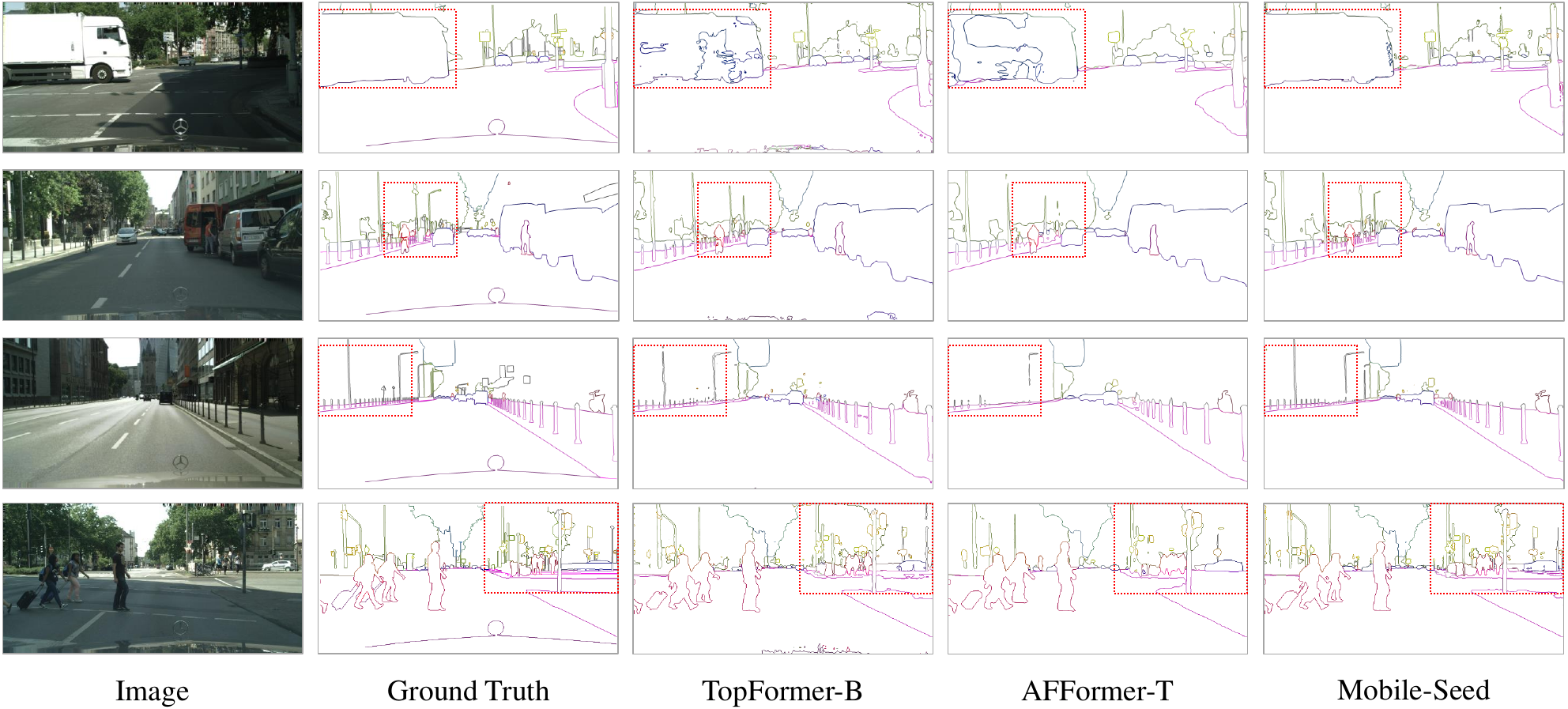}
  \vspace{-0.2cm}
  \caption{Qualitative results of the semantic boundary.}
  \label{fig:sb}
\end{figure*}

To demonstrate that Mobile-Seed achieves a more precise boundary location, we evaluate the semantic boundary accuracy with the F-score metric reported in Tab.~\ref{tab:table3}. It shows that our method outperforms the baseline method by a large margin, especially in the strictest condition with the 3px threshold (about $4.2\%$ improvement in the F-score metric). The results of semantic boundary validate that jointly learning semantic segmentation and boundary detection boosts the segmentation performance in boundary areas. The qualitative results of the semantic boundary in Fig.~\ref{fig:sb} show that our method predicts sharper and more continuous boundaries. 

\begin{figure}[h]
\vspace{-0.4cm}
  \centering
  \includegraphics[width=0.9\linewidth]{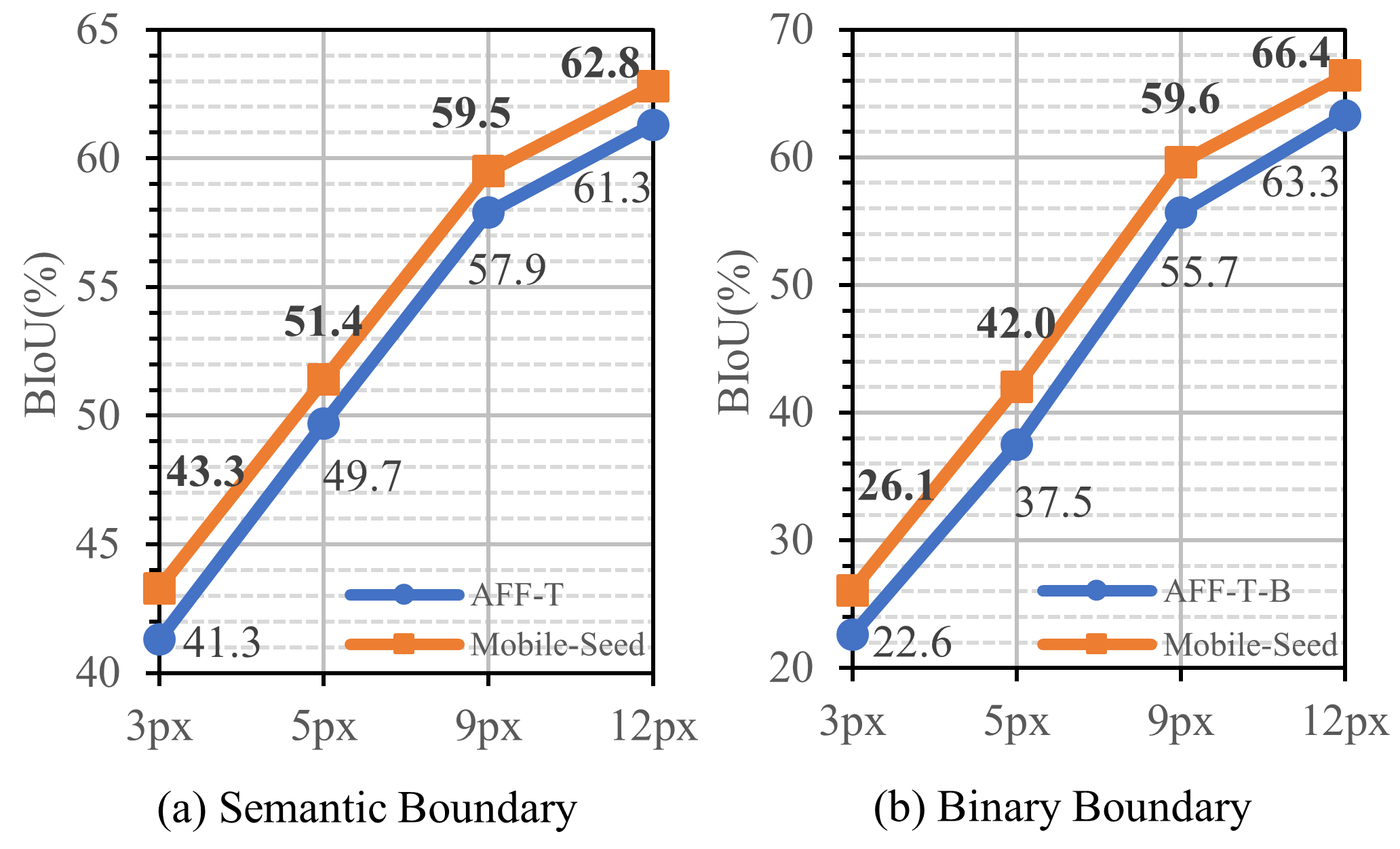}
  \vspace{-0.2cm}
  \caption{ (a) Semantic boundary results and (b) Binary boundary results on the Cityscapes \textit{val} dataset. AFF-T-B means AFFormer-T for the binary boundary detection task. }
  \label{fig:biou}
  % \vspace{-0.4cm}
\end{figure}

Lastly, we report the semantic boundary and binary boundary performance with the BIoU metric. Fig.~\ref{fig:biou} (a) shows the Mobile-Seed achieves higher   BIoU value under several thresholds. As the baseline method is a semantic segmentation framework and has no boundary stream, we retrained it with the binary boundary $\hat{\boldsymbol{b}}$ supervision (called AFF-T-B in the following). The BIoU scores of Mobile-Seed and AFF-T-B shown in Fig.~\ref{fig:biou} (b) demonstrate that our framework extracts crisper and more accurate boundaries than independently learning object boundaries. 

\begin{figure}[ht]
\vspace{-0.2cm}
  \centering
  \includegraphics[width=0.9\linewidth]{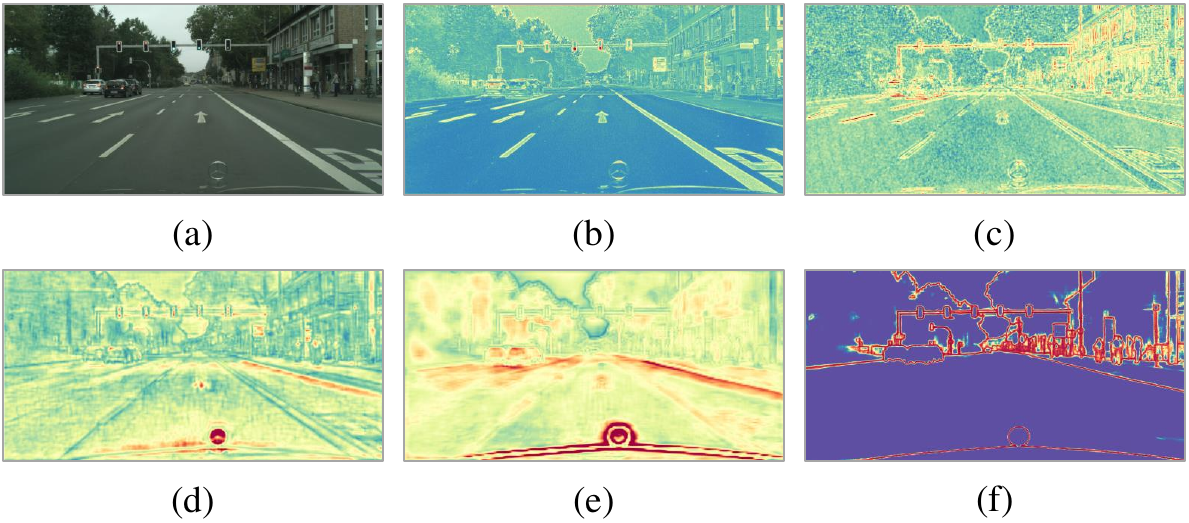}
  \vspace{-0.2cm}
  \caption{Visulization of multi-scale activation maps in boundary stream.  (a) input image. (b)  stage I. (c) stage II. (d) stage III. (e) stage IV. (f) final prediction.}
  \label{fig:edge_activation}
  % \vspace{-0.2cm}
\end{figure}

We additionally visualize the activation maps of each stage from the boundary stream in Fig.~\ref{fig:edge_activation}, illustrating that the lower stages are interested in sharp intensity change (Fig.~\ref{fig:edge_activation} (b), (c)) and higher stages (Fig.~\ref{fig:edge_activation} (d), (e)) focus on semantic inconsistency. Intuitively, the bottom layers capture low-level details and the top layers obtain high-level semantics, and in the end, the boundary stream head adaptively combines multi-level features for boundary prediction.·

\begin{table}[t]
\centering
\footnotesize
\renewcommand\arraystretch{1.1}
\setlength{\tabcolsep}{1.5pt}
\caption{Semantic segmentation results on Cityscapes \textit{val} dataset. LRFormer-T$^\star$: code of LRFormer is not available.}
\label{tab:table1}
\begin{tabular}{l|c|c|c|c}
\hline
Method            & \#Params & FLOPs  & mIoU  & FPS \\ \hline
FCN \cite{long2015fully}              & 9.8M     & 317G   & 61.5  &  11.2   \\  
PSPNet \cite{zhao2017pyramid}          & 13.7M    & 423G   & 70.2  &  9.5   \\ 
DeepLabV3+ \cite{chen2018encoder}      & 15.4M    & 555G   & 75.2  &   8.2  \\ 
SegFormer-B0  \cite{xie2021segformer}        & 3.8M     & 125G   & 76.2  &  11.7   \\ 
TopFormer-B \cite{zhang2022topformer}   &   5.1M       & 11.2G  & 75.2  &   55.6  \\  
PIDNet-S \cite{xu2023pidnet}         & 7.6M     & 47.6G  & 78.7  & 15.3 \\
LRFormer-T$^\star$\cite{wu2023low}        & 13.0M    & 122.0G & 80.7  & -    \\
AFFormer-T\cite{dong2023head}     & 2.2M     & 23.6G  & 76.2  & 27.8 \\
AFFormer-B \cite{dong2023head}    & 3.0M     & 33.5G  & 77.8  & 21.2   \\ \hline
Mobile-Seed(Ours)        & 2.4M   & 31.6G & 78.4 & 23.9   \\ \hline
\end{tabular}
\vspace{-0.4cm}
\end{table}

\begin{table*}[t]
\centering
\footnotesize
\renewcommand\arraystretch{1.1}
\setlength{\tabcolsep}{3.5pt}
\caption{Comparison class-aware semantic segmentation results to the baseline method. AFF-T is short of AFFormer-T.}
\label{tab:table2}
\vspace{-0.1cm}
\begin{tabular}{c|ccccccccccccccccccc|c}
\hline
mIoU       & road           & s.walk       & build       & wall           & fence          & pole           & t-light  & t-sign   & veg     & terrain       & sky            & person         & rider          & car            & truck          & bus            & train          & motor     & bike        & mean           \\ \hline
AFF-T   & 98.2          & 85.3          & 92.5          & 54.7          & 57.6          & 63.9          & 70.1          & 78.5          & 92.7          & \textbf{66.3} & 94.8          & 81.1           & 60.0          & 94.7          & 70.2           & 80.0          & 69.5          & 61.7          & 75.9          & 76.2          \\ \hline
Ours & \textbf{98.3} & \textbf{85.9} & \textbf{92.8} & \textbf{61.8} & \textbf{58.7} & \textbf{66.7} & \textbf{71.6} & \textbf{79.6} & \textbf{92.9} & 65.9         & \textbf{95.1} & \textbf{82.0} & \textbf{61.4} & \textbf{94.9} & \textbf{78.9} & \textbf{85.8} & \textbf{77.9} & \textbf{63.0} & \textbf{77.5} & \textbf{78.4} \\ \hline
\end{tabular}
\end{table*}

\begin{table*}[t]
\centering
\footnotesize
\renewcommand\arraystretch{1.1}
\setlength{\tabcolsep}{3pt}
\caption{Quantatitave results of semantic boundary on the Cityscapes \textit{val} dataset. AFF-T is short of AFFormer-T.}
\label{tab:table3}
\vspace{-0.1cm}
\begin{tabular}{c|cccccccccccccccccccc|c}
\hline
Thrs                  & Method     & road          & s.walk        & build         & wall          & fence         & pole          & t-light       & t-sign        & veg           & terrain & sky           & person        & rider         & car           & truck         & bus           & train         & motor         & bike          & mean           \\ \hline
\multirow{2}{*}{3px}   & AFF-T   &81.3 & 61.0 & 66.1 & 47.8 & 47.4 & 66.1 & 65.5 & 66.9 & 68.2          & 53.6                                         & 78.8 & 57.6 & 65.2 & 74.7 & 77.7 & 85.1 & 91.3 & \textbf{77.5} & 61.2 & 68.0           \\
& Ours & \textbf{84.2} & \textbf{66.8} & \textbf{72.0} & \textbf{57.0} & \textbf{53.4} & \textbf{73.6} & \textbf{68.5} & \textbf{70.1} & \textbf{73.8} & \textbf{59.5}                                         & \textbf{82.2} & \textbf{61.1} & \textbf{66.5} & \textbf{79.3} & \textbf{81.0} & \textbf{88.2} & \textbf{95.2} & 76.8 & \textbf{62.6} & \textbf{72.2}  \\ \hline
\multirow{2}{*}{5px} & AFF-T   & 86.9          & 70.1          & 75.2          & 50.5          & 50.1          & 72.9          & 71.8                               & 74.9          & 78.6          & 57.6                                                  & 85.5          & 65.7          & 69.9          & 82.7          & 78.8          & 86.3          & 91.6 & \textbf{78.8} & 68.7          & 73.5           \\  
& Ours & \textbf{88.6} & \textbf{74.2} & \textbf{80.0} & \textbf{59.4} & \textbf{56.2} & \textbf{78.6} & {\textbf{74.0}} & \textbf{76.4} & \textbf{82.8} & \textbf{63.0}                                         & \textbf{87.9} & \textbf{68.2} & \textbf{71.2} & \textbf{85.7} & \textbf{81.8} & \textbf{89.3} & \textbf{95.6} & 78.4 & \textbf{69.8} & \textbf{76.9}  \\ \hline
\multirow{2}{*}{9px} & AFF-T   & 90.7          & 76.8          & 82.5          & 53.2          & 53.0          & 77.1          & 76.4          & 79.8          & 86.5          & 61.3                                                  & 89.1          & 71.7          & 74.2          & 87.9          & 79.8          & 87.6          & 91.9          & \textbf{80.0}          & 75.8          & 77.6           \\  
& Ours & \textbf{91.5} & \textbf{79.4} & \textbf{86.1} & \textbf{61.8} & \textbf{59.0} & \textbf{82.0} & \textbf{77.4} & \textbf{80.3} & \textbf{89.0} & \textbf{66.3}                                         & \textbf{90.9} & \textbf{73.5} & \textbf{75.4} & \textbf{89.9} & \textbf{82.6} & \textbf{90.3} & \textbf{95.9} & 79.8 & \textbf{75.9} & \textbf{80.4} \\ \hline

\multirow{2}{*}{12px}  & AFF-T   & 91.9          & 79.0          & 85.1          & 54.3          & 54.3          & 78.8          & 77.6         & 81.4          & 89.1          & 62.6                                                  & 90.3          & 73.8          & 75.8          & 89.7          & 80.4          & 88.1          & 92.0          & \textbf{80.6}          & 78.4          & 79.1           \\
& Ours & \textbf{92.4} & \textbf{81.4} & \textbf{88.3} & \textbf{62.9} & \textbf{60.3} & \textbf{83.3} & \textbf{78.4} & \textbf{81.6} & \textbf{91.1} & \textbf{67.6}                                         & \textbf{91.8} & \textbf{75.3} & \textbf{77.0} & \textbf{91.3} & \textbf{83.0} & \textbf{90.6} & \textbf{96.0} & 80.4          & \textbf{78.1} & \textbf{81.6}  \\ \hline
\end{tabular}
\end{table*} 

%%%%%%%%%%%%%%%%%%%%%%%%
\textbf{Extensive validations.} Furthermore, we evaluate our method on the CamVid and PASCAL Context datasets. We retrain the baseline AFFormer-T and report the segmentation and boundary result in terms of IoU and BIoU respectively. Quantitative results in Tab.~\ref{tab:camvid_pascal} show that our method significantly improves semantic segmentation accuracy in various datasets, demonstrating the generalization ability. 

\begin{table}[t]
%\vspace{-0.2cm}
\centering
\footnotesize
\caption{Comparison with baseline method on CamVid and PASCAL Context datasets. PASCAL$^{59}$ and PASCAL$^{60}$ mean PASCAL Context dataset with 59 and 60 categories, respectively. The threshold of BIoU is set to 3px.}
\label{tab:camvid_pascal}
% \vspace{-0.1cm}
\begin{tabular}{ccccccc}
\hline
\multirow{2}{*}{Method} & \multicolumn{2}{c}{CamVid} & \multicolumn{2}{c}{PASCAL$^{59}$} & \multicolumn{2}{c}{PASCAL$^{60}$} \\ \cline{2-7} 
                        & mIoU      & BIoU      & mIoU        & BIoU        & mIoU        & BIoU       \\ \hline
AFFormer-T                   & 71.6      & 41.2           & 45.7        & 20.7             & 41.4        & 14.9             \\
Mobile-Seed                    & \textbf{73.4}      & \textbf{45.2}           & \textbf{47.2}        & \textbf{22.1}             & \textbf{43.0}        &   \textbf{16.2}               \\ \hline
\end{tabular}
\vspace{-0.2cm}
\end{table}

\subsection{Ablation Studies and Insights}\label{sec:ablation}

\textbf{Effectiveness of dual-task learning framework.} We conduct an ablation study to demonstrate thaxt our dual-task learning framework is better than learning semantic segmentation task individually, as shown in Tab.~\ref{tab:dual_task}. This ablation experiment employs the single semantic stream $\mathcal{S}$ as the baseline [A] and test boundary stream $\mathcal{B}$, boundary loss function $\mathcal{L}_{b}$ and dual-task regularization loss $\mathcal{L}_{reg}$, respectively. [B] shows that adding ''multi-scale'' features from the boundary stream boosts the semantic segmentation performance. We do not refer to the ``multi-scale'' features as ''boundary'' features because the boundary supervision is removed. [C] shows that explicitly supervising the boundary stream with the $\mathcal{L}_{b}$ leads to performance degradation, as the mIoU drops about $0.7\%$. This circumstance proves our suppose that applying distinctive supervision to different modules may harm the framework. [D] shows that our dual-task regularization loss $\mathcal{L}_{reg}$ could mitigate the learning divergence and promote the semantic segmentation and boundary detection tasks learning in a complementary way.

\begin{table}[h]
% \vspace{-0.1cm}
\centering
%\normalsize
\footnotesize
\caption{Ablation study on our dual-task learning framework. $\mathcal{S}$ means semantic stream, and $\mathcal{B}$ means boundary stream. $\mathcal{L}_{b}$ means boundary loss and $\mathcal{L}_{reg}$ means dual-task regularization loss. The threshold of BIoU metric is set to 3px.}
\label{tab:dual_task}
\vspace{-0.1cm}
\begin{tabular}{c|cc|cc|cc} \hline
    & $\mathcal{S}$ & $\mathcal{B}$ & $\mathcal{L}_{b}$ & $\mathcal{L}_{reg}$ & mIoU & BIoU  \\ \hline
    $[A]$ & \checkmark &            &            &            & 76.2 & 41.3 \\ 
    $[B]$ & \checkmark & \checkmark &            &            & 77.7 & 42.1 \\
    $[C]$ & \checkmark & \checkmark & \checkmark &            & 76.9 &  41.6 \\
    $[D]$ & \checkmark & \checkmark & \checkmark & \checkmark & \textbf{78.4} & \textbf{43.3} \\ 
    \hline
\end{tabular}
%\vspace{-0.2cm}
\end{table}

\begin{table}[h]
\centering
\vspace{+0.2cm}
%\normalsize
\footnotesize
\caption{Ablation study on feature fusion methods. `ADD' means features addition, `CAT' means features concatenation and `AFD' means active fusion decoder proposed in our method.}
\label{tab:table5}
%\vspace{-0.1cm}
\begin{tabular}{ccc|ccc}
\hline
                  ADD        & CAT        & AFD        & mIoU          & FLOPs & FPS \\ 
    \hline
      \checkmark &            &            & 77.7          & 0.96G & \textbf{24.3} \\
      &           \checkmark &            & 78.0          & 1.91G & 23.6\\
      &              & \checkmark & \textbf{78.4} & \textbf{0.96G}  & 23.9 \\ \hline
\end{tabular}
% \vspace{-0.4cm}
\end{table}

\textbf{Comparison of feature fusion methods.} We conduct ablation studies to prove the effectiveness of our AFD. We take the semantic stream $\mathcal{S}$, boundary stream $\mathcal{B}$ and total loss $\mathcal{L}$ as baseline and check out the influence of feature fusion methods. We compare our AFD  with fixed weight fusion methods, as addition and concatenation in previous approaches. Tab.~\ref{tab:table5} shows that our AFD is more lightweight than concatenation and achieves better performance compared to both addition and concatenation. The results support our assumption that the fusion weights should be conditioned on the input, where our AFD dynamically assigns proper weights to each channel of semantic and boundary features and outperforms addition and concatenation.

%%%%%%%%%%%%%%%%%%%%%%%%%%%%%%%%%%%%%%%%%%%%%%%%%%%%%%%%%%%%%%%%%%%%%%%%%%%%%%%%
\section{Conclusion}\label{sec:6}
\label{sec:conclusion}

In this paper, we present a novel lightweight framework Mobile-Seed for joint semantic segmentation and boundary detection. Our method consists of a two-stream encoder and an active fusion decoder (AFD), where the encoder extracts semantic and boundary features respectively, and the AFD assigns dynamic fusion weights for two kinds of features. Moreover, regularization loss is introduced to alleviate the divergence in dual-task learning. We have implemented and evaluated our approach on various datasets and provided comparisons to other existing techniques. The experimental results validate that our method outperforms all the existing methods and support all claims made in this paper. We believe that the Mobile-Seed can be deployed on lightweight robotics platforms and serves for semantic SLAM, robot manipulation and other downstream tasks.
% \vspace{-0.1cm}

%%%%%%%%%%%%%%%%%%%%%%%%%%%%%%%%%%%%%%%%%%%%%%%%%%%%%%%%%%%%%%%%%%%%%%%%%%%%%%%%
% Only if applicable
%\section*{Acknowledgments}
%We thank XXX for fruitful discussions and for \dots

\bibliographystyle{IEEEtran}

\bibliography{glorified,new}

\end{document}